\documentclass{IEEEcsmag}

\usepackage[colorlinks,urlcolor=blue,linkcolor=blue,citecolor=blue]{hyperref}
\expandafter\def\expandafter\UrlBreaks\expandafter{\UrlBreaks\do\/\do\*\do\-\do\~\do\'\do\"\do\-}
\usepackage{orcidlink}
\usepackage{comment}

\jvol{XX}
\jnum{XX}
\paper{8}
\jmonth{}
\jname{Computing in Science \& Engineering}
\jtitle{Training Next Generation AI Users and Developers at NCSA}
\pubyear{}

\begin{document}

\sptitle{DEPARTMENT: EDUCATION}

\title{Training Next Generation AI Users and Developers at NCSA}

\author{Daniel S. Katz\orcidlink{0000-0001-5934-7525}}
\affil{University of Illinois Urbana-Champaign, Urbana, IL, 61801, USA}

\author{Volodymyr Kindratenko\orcidlink{0000-0002-9336-4756}}
\affil{University of Illinois Urbana-Champaign, Urbana, IL, 61801, USA}

\author{Olena Kindratenko\orcidlink{0009-0002-4280-0260}}
\affil{University of Illinois Urbana-Champaign, Urbana, IL, 61801, USA}

\author{Priyam Mazumdar\orcidlink{0009-0004-8851-5531}}
\affil{University of Illinois Urbana-Champaign, Urbana, IL, 61801, USA}

\markboth{EDUCATION}{EDUCATION}

\begin{abstract} 
This article focuses on training work carried out in artificial intelligence (AI) at the National Center for Supercomputing Applications (NCSA) at the University of Illinois Urbana-Champaign via a research experience for undergraduates (REU) program named FoDOMMaT.
It also describes why we are interested in AI, and concludes by discussing what we've learned from running this program and its predecessor over six years.

\end{abstract}

\maketitle

\chapteri{I}n computationally-focused research organizations, Artificial Intelligence (AI) has become both a tool to be applied and an area of research. The National Center for Supercomputing Applications (NCSA) at the University of Illinois Urbana-Champaign is such an organization, where researchers and cyberinfrastucture (expertise, computing, data, software, networking, security, etc.) have been brought together for over 35 years to work on asking and solving some of the world's most challenging research questions.

In this article, we provide some context about NCSA, describe how AI is an element of our strategy, then talk about our AI-centered research experience for undergraduates (REU) program, FoDOMMaT, including how training happens within it and how it is evaluated, and conclude by discussing what we've learned from our experiences.

\section{NCSA'S CONTEXT}

The University of Illinois Urbana-Champaign is a public US land-grant university, with the mission charged by the state ``to enhance the lives of citizens in Illinois, across the nation and around the world through our leadership in learning, discovery, engagement and economic development.'' \cite{aboutUIUC}

Research is an important part of the university's work: ``At Illinois, our focus on research shapes our identity, permeates our classrooms and fuels our outreach. Fostering discovery and innovation is our fundamental mission. As a public, land-grant university, we have the responsibility to create new knowledge and new ideas and translate these into better ways of working, living and learning for our state, nation and world.'' \cite{aboutUIUC}

At the university, the National Center for Supercomputing Applications, which is one of ten campus interdisciplinary research institutes, has the vision of ``a future enlightened by our research discoveries, where the boundaries of human understanding are continually extended to improve the world,'' and the mission to ``bring people, computing and data together to benefit society.''
As of December 31, 2022, NCSA had 242 staff and postdocs, 184 campus affiliates from 58 departments in 9 colleges and schools, and 143 students participating in staff and faculty affiliate-led research, as well as USD\$52 million of 2022 grant expenditures and an overall project portfolio of USD\$408 million.

At NCSA, as at many other research organizations, \textit{strategy} is a method to balance bottom-up discovery- and curiosity-driven research activities (across many areas) with top-down strategic aims (in a few areas).
Most of NCSA's resources are research grants for specific research, and most of NCSA’s university funding is dedicated to specific support activities, leaving only a small amount of funding that can be used for strategic growth.
Bottom-up successes can lead to changes in strategy, while strategic investments can support individual activities and areas.
Two current strategic foci are quantum computing and artificial intelligence (AI).
We consider these areas in terms of applications, infrastructure, and expertise.
To grow the areas, we both
1) partner with existing applications, infrastructure, and experts on campus and off, and
2) develop our own applications, infrastructure, and expertise.

\section{AI AS AN ELEMENT OF STRATEGY}

To develop our AI area, we've created a Center for AI Innovation (CAII), led by Dr. Volodymyr Kindratenko.

CAII has three themes: research, where it brings the university AI research community together for opportunities to collaborate; scholarship, where it provides students with opportunities to learn and work in AI domain; and industry, where it aligns academic research with industry challenges and opportunities.
Overall, CAII partners with leading researchers and technology developers to bring state-of-the-art AI capabilities to the University research community 

In terms of partnering with researchers, we've won a number of projects where CAII collaborates with other groups, in computer science, high energy physics, life sciences, with partners from Illinois, other universities, national labs, and industry.
And in terms of partnering with technology developers, we've won awards for AI-focused systems:

\begin{itemize}
 
\item HAL is an AI research platform funded by NSF's Major Research Instrumentation (MRI) program, with 16 IBM Power9 servers, each  with 4x NVIDIA V100 GPUs; a DDN parallel file system; and a Mellanox EDR IB interconnect. 
\item Delta is a national production HPC system funded by NSF Innovative HPC program. It's the most performant NSF GPU computing resource, with 124 quad-core SSD CPU nodes, 200 quad-CPU (A100/A40) SSD nodes, a SlingShot network, and 10 PB Lustre \& Flash storage.  
\item DeltaAI is a national production HPC system funded by NSF Innovative HPC program, currently being built. It will be an advanced computing and data resource (``a vast array of next-generation GPUs'') as a companion system to Delta, and will triple NCSA’s AI-focused computing capacity.  

\end{itemize}

And building on previous student programs, we won an NSF Research Experiences for Undergraduates (REU) award called FoDOMMaT: the Future of Discovery: Training Students to Build and Apply Open Source Machine Learning Models and Tools. V. Kindratenko is PI, D. S. Katz is co-PI, and O. Kindratenko is program coordinator.

\section{FoDOMMaT}

FoDOMMaT~\cite{FoDOMMaT} is a summer REU program, that runs for 10 weeks on-site at NCSA. Students work with pairs of mentors, typically one of who has expertise in a research discipline where AI/ML can be used, while the other has expertise in AI/ML. Students also can work in pairs, if the research project can be broken down in such a manner. In our previous REU program (INCLUSION~\cite{INCLUSION}), we experimented with the pairs of students having differing levels of expertise so that one could get experience in teaching/mentoring the other, and the other might be able to do the same thing in a future year.

We developed a comprehensive application and review process to select students for the program. It's open to all majors and all years of study, though some software development experience with Python is required, as well as some exposure to machine learning via coursework, self-study, or other projects. The application process opens in December and closes in March, and in it, students are asked to explain why they are interested in this specific REU program and how they would benefit from it, as well as to submit a short statement of purpose, resume, and an unofficial transcript. They also have to choose 2-3 projects from the list of offered projects that they are most interested in. Their applications are holistically reviewed by the program coordinator and the project mentors on the rolling basis, including an on-line interview, and offers are sent as soon as a qualified student is identified. Reference letters are requested at this time as well. 

As with most NSF-funded summer REU programs, students receive a \$600/week stipend, paid housing, a meal plan, and one round trip to/from campus.
The program seeks to bring in ten undergraduate students each summer, to give them opportunities to learn and contribute to AI research, focusing on those who would otherwise lack such an opportunity, to work on cutting-edge and socially impactful projects. The students also benefit from the training and the research experience by potentially considering graduate school and being more ready for their career, perhaps in industry.
It also allows mentors to explore new research ideas, and to potentially find new graduate student.
Finally, the projects create tools and products that can benefit the scholarly community.

The first week of the program prioritizes orientation, team building (of both the overall cohort of students and the research teams, and training on machine learning (ML) methods.
The remaining nine weeks are when the students work on their specific research projects. 
Additionally, during this time, the students attend seminars by visiting researchers from industry and academia, social activities, networking and professional development events.
By the end of the program, the students will have openly shared models and tools, including in an annual summer undergraduate research symposium that is university-wide, and in the research community associated with their discipline, and ideally, at their home institution via their undergraduate advisor.

Specific students activities include:

\begin{itemize}
\item Write a research plan (2 pages) early in the program
\item Attend scheduled joint lightning talks with other undergraduate programs
\item Give one lightning talk on the progress of their project
\item Attend scheduled FoDOMMaT social and professional development events 
\item Meet regularly with their FoDOMMaT mentors and NCSA Senior Research Coordinator
\item Write a research report (3--4 pages)
\item Write weekly short reports about project work and progress
\item Participate in NCSA Summer Research Poster Session and Illinois Summer Undergraduate Research Symposium
\end{itemize}

NCSA also offers other student internship programs, such as Students Pushing INnovation (SPIN), which  supports students from the University of Illinois for both academic year and summer sessions, and an International Research Internship (IRI) program. During the summer, students from these three programs share the same student space, some of the activities, including social events and lightning talks, and have a chance to interact and learn from each other. Occasionally, they also contribute to the same projects. We and other REU programs on campus have also formed the Illinois Summer Research Program Alliance to organize joint events for students across, mostly centered on professional development activities and graduate school application. At the end of the term, a joint Summer Undergraduate Research Symposium is organized with several participating programs, including FoDOMMaT, SPIN, and IRI. 

\subsection{TRAINING AND PROJECTS}

FoDOMMaT's initial training is structured to bring all participants to the same level of familiarity with the relevant tools, technologies, and methodologies around developing open-source code and ML models. It's conducted by a graduate student program mentor who continues to work with the cohort for the rest of the program as an ML consultant. On the first day, we provide students with access to HAL, the computer system with GPU resources used for the projects throughout the program, and review basics of Linux, SSH, Bash, git, Python, and good code development practices. We also try to highlight the importance of good data management as well as some introduction to data-driven storytelling and visualization as these skills are necessary for any research project. 

We next review the intuition of what different machine learning methods are attempting to do and how they can be leveraged to solve real problems. The main distinction we draw are differences between supervised and unsupervised methods, allowing us to explore and implement examples on linear regression, logistic regression, support vector machines, random forests, gradient boosting and different clustering algorithms. Another area we focus on are different data compression techniques such as PCA and T-SNE as a tool to manage high dimensional data. This provides a large survey of methods that the students can potentially try for their own projects. 

We then move from standard machine learning into deep learning with PyTorch. To extend their knowledge from machine learning and to illustrate how deep learning is a logical next step for more complex learning algorithms, we reimplement linear and logistic regression using PyTorch. This gives an opportunity to understand all the pieces that go into deep learning: (1) Visualize how we utilize backpropagation and gradient descent for optimization, (2) Demonstrate how to prepare and batch data for training, (3) How to select a loss function that matches our task, (4) Understanding the benefits of different optimizers and (5) How activation functions enable non-linear learning in our model. Towards the end of the first week, students are introduced to more advanced models, such as convolutional and LSTM neural networks, as well as modern techniques and tools to train them such as the HuggingFace platform. We also highlight and implement some examples of Autoencoders to explore how deep learning can be used for data compression and unsupervised learning. Lastly, we focus on some of the more logistical parts of training models such as model debugging, gradient checking, and distributed deep learning for multi-gpu training. 

Throughout the rest of the program, we organize weekly reading groups, where the students are given a list of areas of interest, and we curate papers that match them. We also put a focus on the reimplementation of some of these papers so students can understand exactly how these models are made from scratch. We always begin our sequence of papers with AlexNet as it is a simple model that we implemented previously when learning about convolutional neural networks and it allows us to reference the code as we review the details of the paper. The remaining topics are typically about ResNet, Reinforcement Learning, Generative Diffusion, Transformers, BERT, Vision Transformers, and GPT. We also did a complete re-implementation from scratch and training of ResNet to understand residual connections, Deep-Q Learning on OpenAI Gym games, Variational AutoEncoders as a generative model, the Attention mechanism in Transformers, as well as encoder Transformers applied to images in the Vision Transformers and decoder Transformers applied to natural language in GPT. This provides a robust understanding of not only the papers themselves, but an implementation that can give context for how these models work and new ideas in the future. 

In 2022 and 2023, we offered ten projects each year for the students to choose from. The projects were provided by NCSA staff and faculty affiliates, and they included developing open-source code and ML models. All projects were supervised by two staff/faculty mentors and also involved graduate students from their respective research groups. A description of all the projects is available on-line~\cite{FoDOMMaT}. They range from studying the underpinnings of racial health disparities to digital agriculture to developing deep learning models for multi-messenger astrophysics. In 2024, we are offering seven projects, of which some continue from the previous year and some are new.

\subsection{EVALUATION}

Evaluation is an integral component of the FoDOMMAT project, intended to provide useful and valid information about the program's implementation, effectiveness, recruitment/diversity, outcomes, and sustainability. Within this, its focus is on the effectiveness of activities, the extent to which students have learned/been trained in ML/DL, how their perceived skills with ML/DL have changed as a result of participating in the REU program, and their satisfaction with different aspects of the program. It is conducted on a formative and summative basis. 

The evaluation is conducted by external evaluators, Prof. Cherie Avent and her graduate student Katie Regelson from the Department of Educational Psychology at the University of Illinois. It consists of an initial and final survey conducted in Qualtrics and a focus group that meets at the end of the summer. Results are gathered and analyzed, and findings are shared with the program team in late fall. The evaluation team prepares a detailed report and recommendations that we use for the next term of the program. 

\section{LESSONS AND CONCLUSIONS}

During our six years of running REU programs, we've learned that having a motivated graduate student program mentor involved has been a key element. Such a person understands undergraduates, understands research, and understands mentors. They sit between them all, and make the program and projects work.
In 2022, this role was filled by Priyam Mazumdar, who was a MS student in Statistics, and is now a PhD student in Electrical and Computer Engineering at the University of Illinois Urbana-Champaign.

In addition, the group nature of the REU programs has been a strong contributor to their success. This includes the annual group of students forming a cohort, based on shared training and shared work and social experiences, along with the research groups of formed by the mentors, their groups, and the REU students working together on a research project.

FoDOMMaT has been successful so far, in multiple ways.
We have brought in members of underrepresented groups:
in 2022, five of eight participants were female, two of whom were also minorities, and in 2023, seven of ten students were from underrepresented groups (five female, one Black, one Latino.)
The students have also had successes outside of their completing their summer work: in 2022 one published a conference paper~\cite{REU2022} as first author, and in 
2023, two students were the first two authors of a workshop talk~\cite{LLM-workflow}, three students continued work on two projects in the Fall semester, and two students from underrepresented groups have applied for the University of Illinois's Computer Science PhD program. Our most recent evaluation report (2023) indicates strong student satisfaction on elements such as their interaction with other students, their interaction with their project mentors, and their interaction with the graduate student program mentor.
In addition, we won the 2022 HPCwire Editors' Choice Workforce Diversity \& Inclusion Leadership Award~\cite{HPCWire2022}.

\section{ACKNOWLEDGMENTS}
The authors would like to thank NCSA and the University of Illinois for supporting work in this field. This work was supported by NSF under award \#2050195.

\def\refname{REFERENCES}

\vspace*{-8pt}

\begin{IEEEbiography}{Daniel S. Katz}{\,}is Chief Scientist at the National Center for Supercomputing Applications, and a Research Associate Professor in computer science, electrical and computer engineering, and information sciences with the University of Illinois Urbana-Champaign, Urbana, IL, USA. He received the Ph.D. degree in electrical engineering from Northwestern University, Evanston, IL, USA. His research interests include software applications, algorithms, fault tolerance, and programming in parallel and distributed computing, and policy issues, including citation and credit mechanisms and practices associated with software and data, organization and community practices for collaboration, and career paths for computing researchers. He is a senior member of the IEEE, the IEEE Computer Society, and ACM, co-founder and current Associate Editor-in-Chief of the Journal of Open Source Software, co-founder of the US Research Software Engineer Association (US-RSE), and co-founder and steering committee chair of the Research Software Alliance (ReSA). Contact him at d.katz@ieee.org.
\end{IEEEbiography}

\begin{IEEEbiography}{Volodymyr Kindratenko}{\,}is an Assistant Director at the National Center for Supercomputing Applications, where he leads the Center for AI Innovation, a Research Associate Professor in computer science, and Adjunct Associate Professor in electrical and computer engineering with the University of Illinois Urbana-Champaign, Urbana, IL, USA. He received the D.Sc. degree in analytical chemistry from the University of Antwerp, Belgium. His research interests include high-performance computing, special-purpose computing architectures, cloud computing, and machine learning systems and applications. He is a senior member of the IEEE and ACM and currently serves as an Associate Editor of the International Journal of Reconfigurable Computing. Contact him at kindrtnk@illinois.edu.
\end{IEEEbiography}

\begin{IEEEbiography}{Olena Kindratenko}{\,}is a Senior Research Coordinator at the National Center for Supercomputing Applications, where she is responsible for overseeing and running Center's student internship and faculty affiliates programs and activities, including SPIN and FoDOMMaT. She received the M.Ed. degree in Education Policy, Organization \& Leadership from the University of Illinois Urbana-Champaign, Urbana, IL, USA. Contact her at kindrat2@illinois.edu.
\end{IEEEbiography}

\begin{IEEEbiography}{Priyam Mazumdar}{\,}is a PhD student in the department of Electrical and Computer Engineering at the University of Illinois Urbana-Champaign where he works on low-resource learning for Automatic Speech Recognition. He also actively works with the National Center for Supercomputing Applications on research for Geospatial AI as well as different education initiatives. He has helped build an introductory course to get new students working in Machine Learning and has mentored them throughout the summer REU program to promote good research practices and experiential learning. Contact him at priyamm2@illinois.edu
\end{IEEEbiography}

\end{document}